\documentclass[letterpaper, 10 pt, conference]{ieeeconf}  

\IEEEoverridecommandlockouts                              

\usepackage{graphicx} 
\usepackage{mathptmx} 
\usepackage{times} 
\usepackage{amsmath} 
\usepackage{amssymb}  
\usepackage{hyperref}

\title{\LARGE \bf
Low-Fidelity Visuo-Tactile Pre-Training Improves Vision-Only Manipulation Performance
}

\author{Selam Gano$^{1}$, Abraham George$^{2}$ and Amir Barati Farimani$^{3}$
\thanks{$^{1}$Selam Gano is with the Department of Mechanical Engineering, 
        Carnegie Mellon University, Pittsburgh, PA, United States
        {\tt\small selamg@andrew.cmu.edu}}%
\thanks{$^{2}$Abraham George is with the Department of Mechanical Engineering, 
        Carnegie Mellon University, Pittsburgh, PA, United States
        {\tt\small aigeorge@andrew.cmu.edu}}%
\thanks{$^{3}$Amir Barati Farimani is with Faculty of Mechanical Engineering, 
        Carnegie Mellon University, Pittsburgh, PA, United States
        {\tt\small barati@cmu.edu}}%
}
\begin{document}

\maketitle
\thispagestyle{empty}
\pagestyle{empty}

\begin{abstract}
Tactile perception is essential for real-world manipulation tasks, yet the high cost and fragility of tactile sensors can limit their practicality. In this work, we explore BeadSight (a low-cost, open-source tactile sensor) alongside a tactile pre-training approach, an alternative method to precise, pre-calibrated sensors. By pre-training with the tactile sensor and then disabling it during downstream tasks, we aim to enhance robustness and reduce costs in manipulation systems. We investigate whether tactile pre-training, even with a low-fidelity sensor like BeadSight, can improve the performance of an imitation learning agent on complex manipulation tasks. Through visuo-tactile pre-training on both similar and dissimilar tasks, we analyze its impact on a longer-horizon downstream task. Our experiments show that visuo-tactile pre-training improved performance on a USB cable plugging task by up to 65\% with vision-only inference. Additionally, on a longer-horizon drawer pick-and-place task, pre-training — whether on a similar, dissimilar, or identical task — consistently improved performance, highlighting the potential for a large-scale visuo-tactile pre-trained encoder. Code for this project is available at: \href{https://github.com/selamie/beadsight}{https://github.com/selamie/beadsight}.
\end{abstract}

\section{INTRODUCTION}

\begin{figure}[h]
      \includegraphics[width=\linewidth]{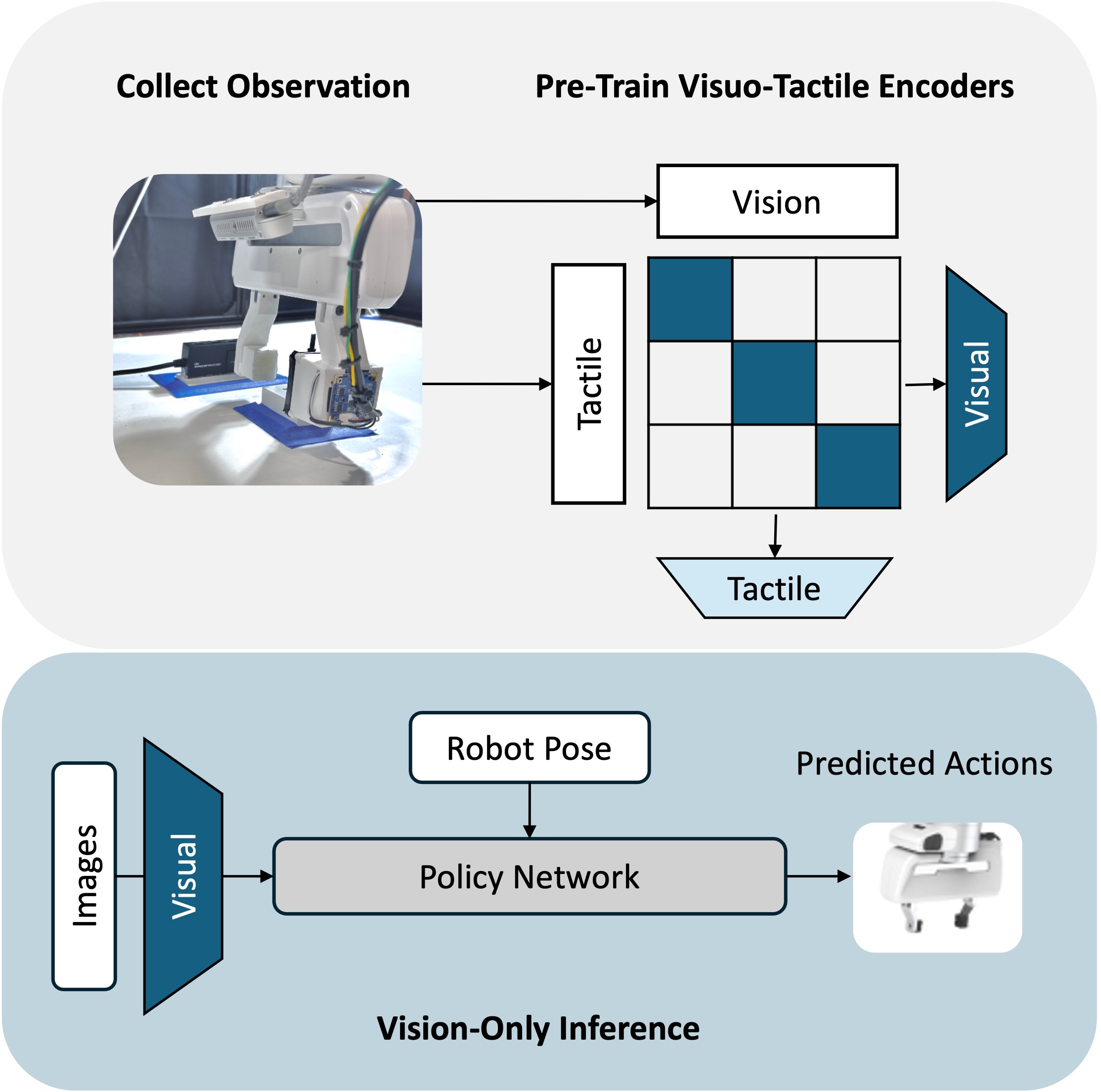}
      \caption{A diagram of the flow of information through tactile pre-training and inference. During contrastive pre-training, primed tactile and visual encoders are produced. The encoders are further refined on the downstream task. Then, the tactile encoder is discarded for vision-only inference without the need for a tactile sensor.}
      \label{fig:fig1}
\end{figure}

Translating advances in visual perception to robotic grasping and manipulation of objects remains challenging. For complex manipulation tasks such as peg insertion, pulling or twisting with resistance, and dynamic motions such as throwing and catching, fine-grained manipulation requires tactile perception. Tactile sensors have been paired with visual sensors for both classical control and machine learning approaches to these tasks \cite{li2020review}, but fragility and cost issues present barriers to heavy use or industrial integration, particularly for manipulation tasks that would place higher forces on these sensors. 

Previously, a GelSight \cite{maldonado2012improving} tactile sensor was used to train an agent on a USB insertion task \cite{george2024visuo}. GelSight is not designed for robustness to higher shear forces and was noted to break irrecoverably during data collection and inference for that task, requiring repeated replacement. This work also demonstrated an approach using tactile information only during pre-training, then disabling the tactile sensor for fine-tuning and inference, achieving a more robust vision-only manipulation system.

BeadSight \cite{george2024beadsight} aimed to make a simpler low-cost, high-durability sensor that, like GelSight, still operated at an end effector's point of contact with objects. We constructed the BeadSight sensor, which does not rely on any calibration and instead relies entirely on neural networks to distill information about contacts and movements on the tactile surface. 

In this work, we repeated the visuo-tactile pre-training USB plugging experiment using the low-fidelity BeadSight sensor to directly compare the effectiveness of the different tactile sensors for training imitation learning agents on dexterous manipulation tasks. We also validate this approach with a second drawer pick-and-place experiment and investigate how using different datasets for visuo-tactile pre-training affects the agent's performance on downstream tasks. Additionally, we confirm that tactile information is responsible for the improvement in agent performance by comparing with a vision-only pre-training approach. 

Our contributions are: 

    \begin{enumerate}
        \item We demonstrate the effectiveness of visuo-tactile pre-training with a low-fidelity, open-source sensor. 
        \item We demonstrate an affordable, durable approach to visuo-tactile pre-training, which is especially cost-effective for improving the performance of vision-only agents.
        \item We show that by freezing a pre-trained tactile encoder or removing it entirely, we can mitigate the overfitting of low-fidelity tactile sensors while still maintaining a performance boost from tactile information.
        \item We demonstrate the unique impact of tactile sensing on performance by comparing with a vision-only pre-training ablation. 
    \end{enumerate}
    
\section{BACKGROUND AND RELATED WORK}
\subsection{Tactile Sensing} 
In robotic manipulation, there are a wide variety of tactile sensors and sensing approaches \cite{li2020review}. These sensors use different modalities to collect tactile information, including forces and torques applied normally or tangentially \cite{choi2010polymer} \cite{schurmann2011modular},  mechanical vibrations \cite{meier2016distinguishing} \cite{molchanov2016contact}, thermal measurement and conductivity \cite{siegel1986integrated} \cite{wade2015handheld}, and pretouch proximity \cite{koyama2018high}. 

Tactile image processing with machine learning and traditional image-based approaches have been used successfully with visuo-tactile sensors that use a camera near the point of contact, with most tracking the deformation of some surface \cite{li2020review}. This paper will focus on the visuo-tactile category of sensors and approaches, which include cameras placed directly on robot fingertips \cite{maldonado2012improving}, and sensors like DenseTact \cite{do2022densetact} \cite{do2023densetact}, See-Through-Your-Skin \cite{hogan2021seeing}, and GelSight \cite{yuan2017gelsight}.

The GelSight sensor is designed to measure geometry with a very high spatial resolution, such that the deformation of its gel-elastomer surface corresponds directly to the exact object shape and tension on the contact surface \cite{yuan2017gelsight}. Markers printed on the gel surface are tracked by an embedded camera, which can estimate shear force and slip state. Deriving exact measurements from GelSight requires an initial calibration step. Previous work using a GelSight Mini for the same task of plugging in a USB cable noted the fragility of the elastomer gel and the sensor itself for tasks that apply higher forces to the contact surface \cite{george2024visuo}.

BeadSight is a low-cost open-source visuo-tactile sensor that uses hydrogel beads as an inexpensive, durable sensing medium. An embedded camera observes the beads' motion and deformation, however, the beads' movements are stochastic and, unlike GelSight, the relationship between the beads and shear forces are not directly computed from first-principle physics methods. Previously, a deep neural network was used to reconstruct contact pressure maps with BeadSight as an example tactile learning approach \cite{george2024beadsight}.    

\subsection{Tactile Sensing and Manipulation Control Policies}
Prior work integrating tactile sensors in control policies range from classical control approaches to reinforcement and imitation learning. \cite{wilson2023cable} used classical state estimation and a heuristic method to follow cables and insert wires. \cite{she2021cable} combined a linear-quadratic regulator (LQR) controller and linear dynamics model to follow an audio cable and successfully plug in a headphone jack. Reinforcement Learning (RL) agents have been trained to perform peg insertion, door opening, and in-hand rotation tasks \cite{chen2022visuo, sferrazza2023power}.  However, these approaches require simulation for the sheer amount of training required ($\sim$ 1 million steps). Nearest Neighbors Imitation Learning \cite{pari2021surprising} is a method that encodes observations and demonstrations into a latent space and compares the distance between them; \cite{yu2023mimictouch} used this method with online residual reinforcement learning to reduce computation intensity and learn visuo-tactile peg insertion. \cite{guzey2023dexterity} used a similar approach to learn a variety of tasks including bowl and cup unstacking, bottle opening, and joystick movement. 

\subsection{Imitation Learning Methods}
Imitation Learning (IL) methods aim to mimic human behavior by learning from human demonstrations, which can be easier than designing optimal reward functions as in reinforcement learning \cite{sasaki2020behavioral}.  Variations of this method have trained manipulation agents to perform kitchen tasks and stack blocks \cite{shafiullah2022behavior, george2023one} and even sculpt clay \cite{bartsch2024sculptdiff}. IL tends to perform more poorly for more complex action sequences, particularly with non-deterministic goal policies and significant shifts between training and deployment domains. A state-of-the-art method was selected that aims to address these drawbacks, Diffusion Policy \cite{chi2023diffusion}, which uses diffusion to better model multi-modal action pathways. 

\subsubsection{Diffusion Policy} Prior work in \cite{chi2023diffusion} constructs a diffusion framework that generates observation-conditioned action sequences,  $p(\mathbf{O}_t | \mathbf{A}_t)$, where action sequence $\mathbf{A}_t$ is conditioned on $\mathbf{O}_t$, an observation at timestep $t$. A sample of examples $\mathbf{A}_t^0$ is drawn from the dataset during training, and a random noise $\epsilon^k$ is sampled for denoising step $k$. The noise prediction network $\epsilon_\theta$ predicts noise from the noised example data using the loss function: 

$$Loss = MSE(\epsilon^k,\epsilon_\theta(\mathbf{O}_t,\mathbf{A}_t^0 +  \epsilon^k, k)) $$

The denoising diffusion model \cite{ho2020denoising} iteratively denoises a Gaussian sample representing the action sequence:

$$\mathbf{A}_t^{k-1} = \alpha (\mathbf{A}_t^k - \gamma \epsilon_\theta(\mathbf{O}_t,\mathbf{A}_t^k, k) + \mathcal{N}(0,\sigma^2I))$$

Feature-Wise Linear Modulation (FiLM) layers in the noise prediction network enable a conditioning encoder to influence the network \cite{perez2018film}; observation $\mathbf{O}_t$ is used to condition the denoising process this way. 

\subsubsection{Action Chunking Transformers (ACT)}
ACT trains a Conditional Variational Autoencoder (CVAE) combined with a transformer backbone to predict action sequences conditioned on state and vision observations \cite{zhao2023learning}. During training, the latent variable of the CVAE helps the model capture multi-modal policies, while a large KL divergence loss keeps the network from becoming too reliant on the latent information. During inference, the latent variable is set to zero. The autoencoder uses a latent variable to reduce the impact of multi-modal training data, and temporal ensembling at inference reduces the effect of a single bad prediction. Specifically, ACT predicts the goal states $A_{t, t}$ to $A_{t, t+h}$ at each timestep. During inference, all predictions $A_{t-h, t}$ to $A_{t, t}$ of the goal action at the current timestep are combined using a weighted average of exponentially decaying weights, $w_i = e^{-ki}$, and the ensembled action is executed. 

\subsection{Contrastive Pre-training}
Contrastive pre-training methods train models to distinguish between similar and dissimilar data, coding them as positive (similar) or negative (dissimilar) pairs \cite{rethmeier2023primer}. Those modeled after the structure of Contrastive Language Image Pre-training (CLIP) \cite{clipradford2021} do this by maximizing the cosine similarity of the paired data in some multimodal embedding space, while minimizing the similarity of dissimilar pairs. For visuo-tactile robotic manipulation, tactile and visual observations can be compared in that latent space. \cite{kerr2022self} used this approach to better identify flaws in fabric, and \cite{dave2024multimodal, zambelli2021learning} both improved visuo-tactile object identification this way.

\section{METHODS}
\label{sec:methods}

\begin{figure}[h]
      \includegraphics[width=\linewidth]{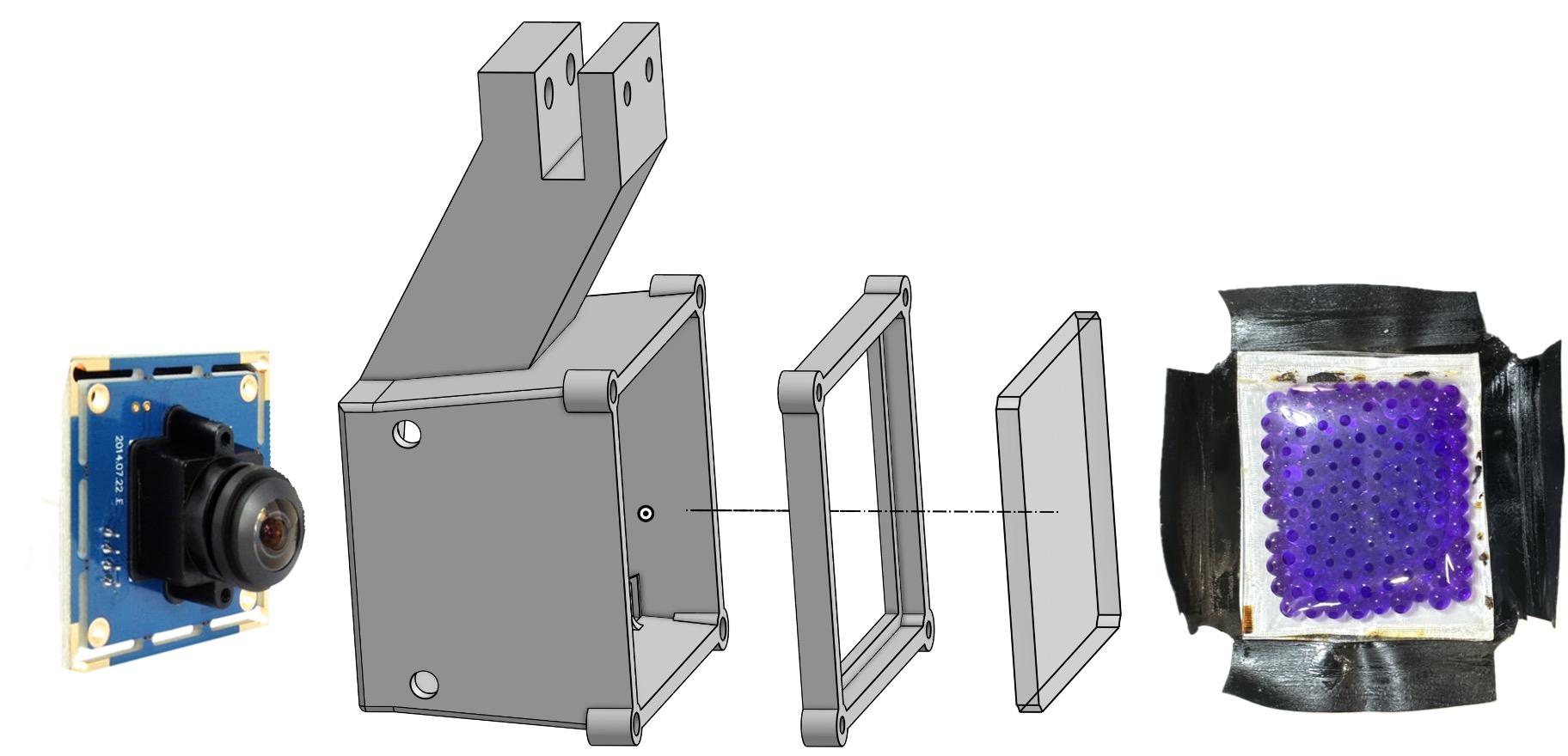}
      \caption{BeadSight components and construction, including (from left to right) the camera, housing, lid frame, acrylic panel, and bead sac on a light-blocking backing }
      \label{fig:beadassembly}
\end{figure}

\begin{figure}[h]
      \includegraphics[width=\linewidth]{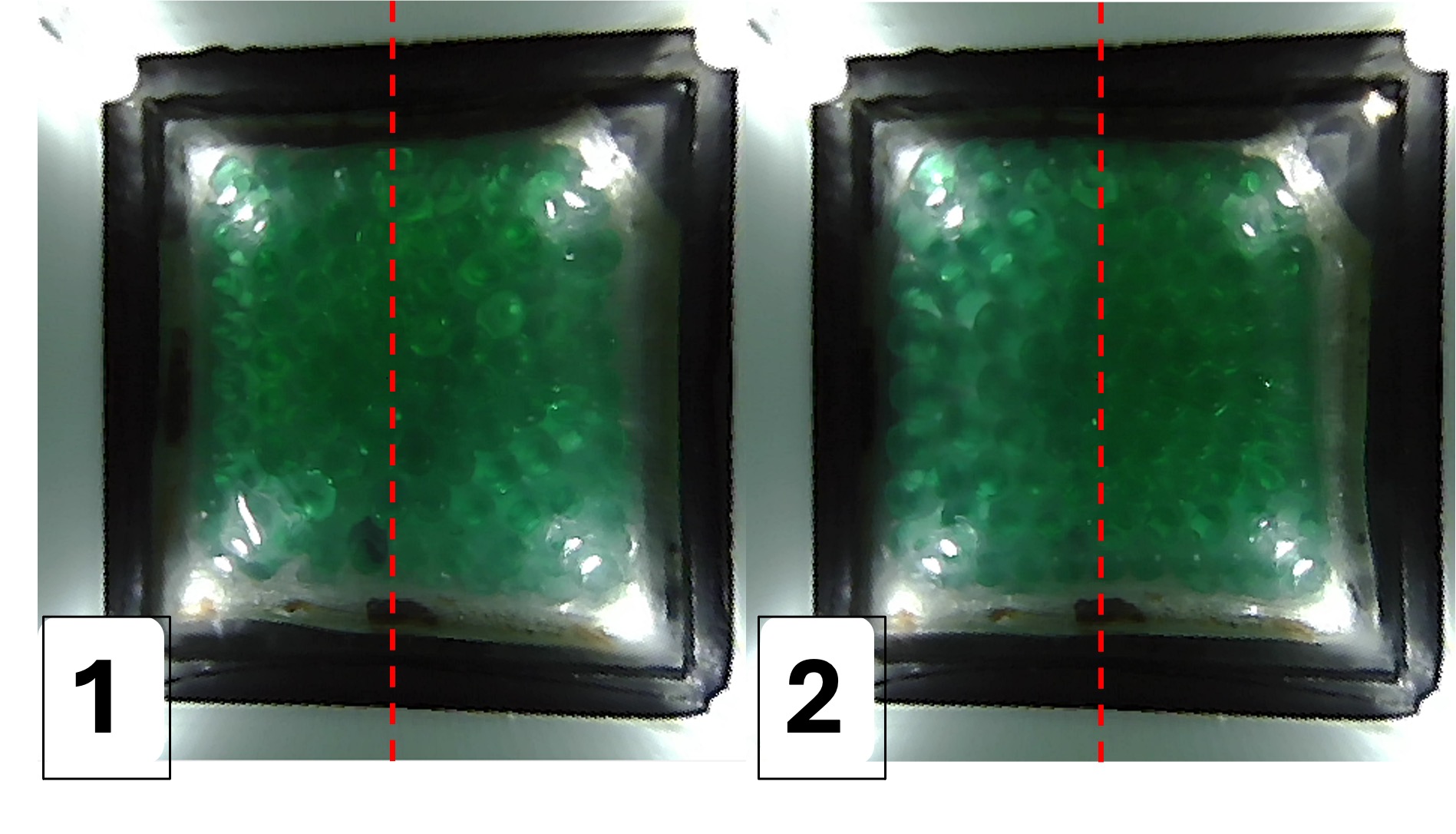}
      \caption{The embedded camera's view of the BeadSight hydro-gel bead sac, relaxed (1) and pressed (2), after down-scaling and correcting fish eye distortion.}
      \label{fig:beadsacs}
\end{figure}

\subsection{BeadSight Hardware}
The BeadSight sensor used in this work measures contact forces using the deformation of hydro-gel beads, captured by a rear-mounted camera. Its 3D-printed enclosure (Figure \ref{fig:beadassembly}) includes threaded heat inserts for attachments and a mount for the Franka Emika Panda robot gripper. The housing and lid allow easy replacement of the contact surface, enhancing durability. The lid consists of a 3D-printed frame, a press-fit acrylic insert, and a hydro-gel bead sac, which is backed with white adhesive tape to improve light reflection. Black PVC adhesive tape seals the sensor from external light sources (Figure \ref{fig:beadsacs}).

Using the same method as \cite{george2024beadsight}, the hydro-gel bead sacs use Polyacrylamide (PAM) beads hydrated with water and sealed in clear PVC sheeting. For this work, each sac contained 50 green PAM beads hydrated with 5ml of water. The embedded camera, also from \cite{george2024beadsight}, is a 1080 × 1920 USB camera with a 180° fish-eye lens, capturing at 30Hz. Four LEDs illuminate the bead sac, and images are downscaled to 480 × 480 resolution with fish-eye correction before processing.

\subsection{Data Collection}
\label{sec:data_collection}

We selected two primary tasks to evaluate the low-fidelity sensor and pre-training method, using the same observation sensors in both cases. For each task, 100 expert demonstrations were collected via teleoperation of a Franka Emika Panda robot. Teleoperation was performed using the Oculus VR method from \cite{george2023openvr}, where handheld Oculus controllers controlled the robot’s end-effector, while the operator viewed the workspace directly rather than through the VR headset.

\begin{figure}[h]
      \centering
      \includegraphics[width=\linewidth]{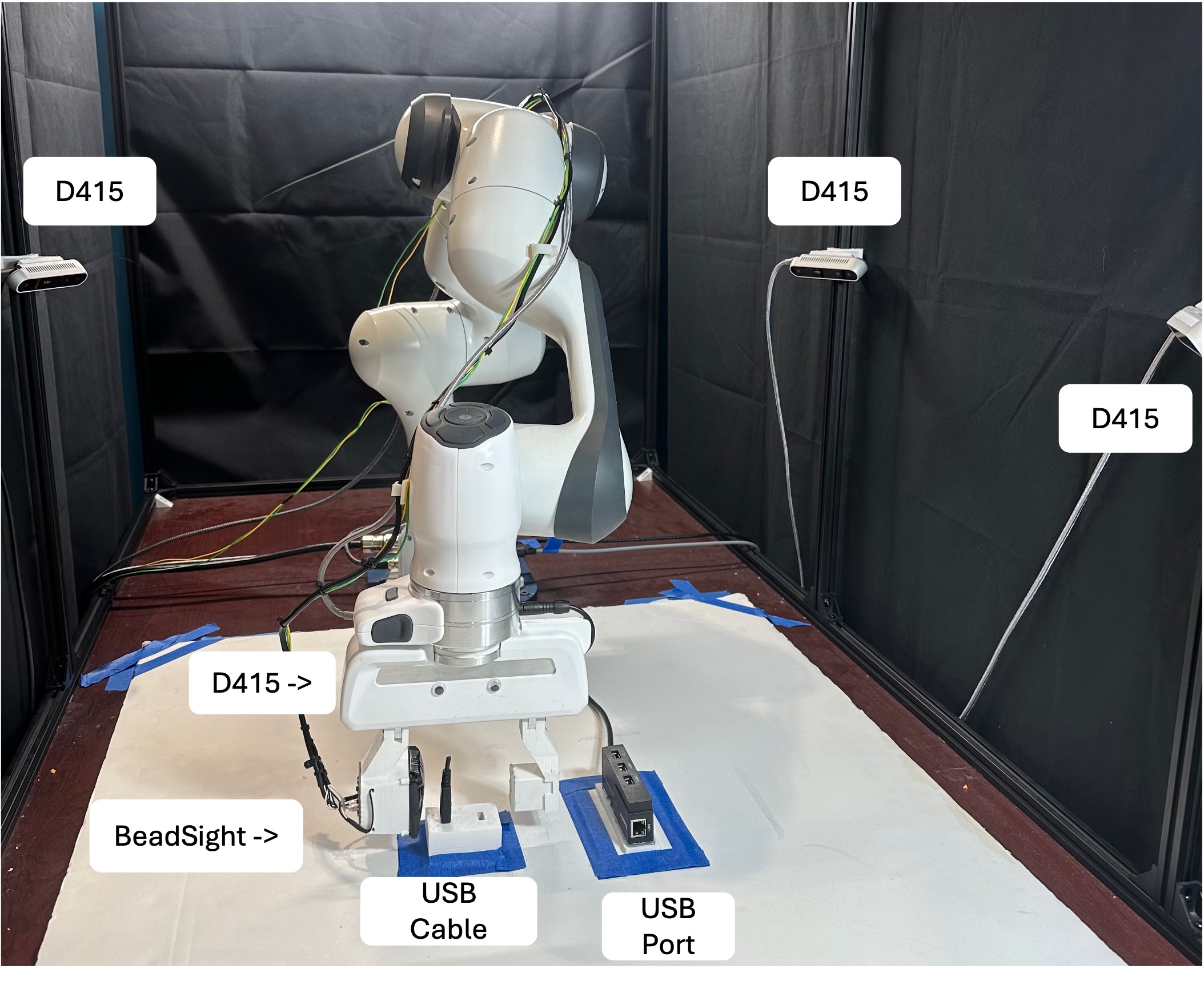}
      \caption{USB cable plugging experimental scene, with a D415 wrist-mounted camera (not visible) positioned behind the robot gripper.}
      \label{fig:usb_scene}
\end{figure}

Visual observations were captured by six Intel RealSense cameras: four D415 cameras around the workspace, a D415 wrist-mounted camera, and a D445 overview camera (Figure \ref{fig:usb_scene}). The BeadSight sensor on the robot's right finger collected tactile data, with a 3D-printed left finger featuring a silicone gel surface to aid grip.

\begin{figure}[h]
      \centering
      \includegraphics[width=\linewidth]{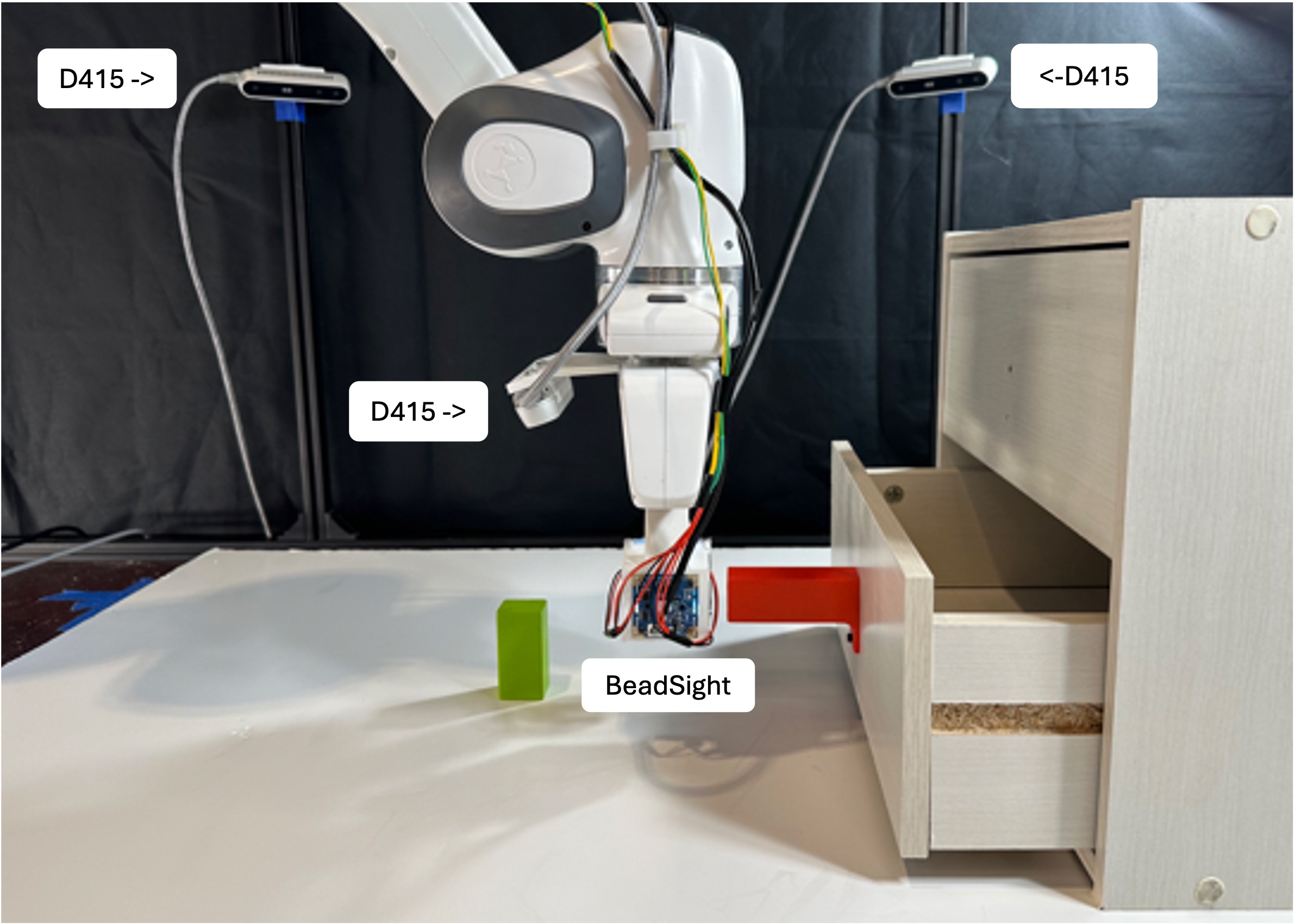}
      \caption{Drawer placement task experimental scene, including a side view of the BeadSight Sensor attached to the end effector.}
      \label{fig:drawer_scene}
\end{figure}

The first task, USB cable plugging (Figure \ref{fig:usb_scene}), uses the same experimental setup as \cite{george2024visuo} to allow for comparison with the GelSight sensor. This dexterous, precision-based task relies heavily on tactile information. For Diffusion Policy, we tested both ablating and freezing the tactile encoder during the downstream task.

The second task is a longer-horizon sequence: opening a drawer, picking up a block, placing it inside, and then closing the drawer (Figure \ref{fig:drawer_scene}). Additionally, we introduced a supportive pre-training task (picking and placing the same block onto the drawer handle) to assess the impact of pre-training with identical, similar, and dissimilar tasks on downstream performance.

\subsection{Tactile Encoder}

Because the BeadSight sensor's hydro-gel beads are not fixed, tactile events are inferred from the beads' relative motion and deformation over time. To capture this relative motion, we introduce a ``tactile horizon", the total number of past tactile frames (including the current frame) combined to form the tactile observation. We use a tactile horizon of 5 frames (the current time step and 4 previous time steps). We collapse the 5 frames on the channel dimension, converting 5 individual $3\times 480 \times480$ images to a single $15\times 480 \times480$ tensor. For a different tactile horizon, the representative tensor would be of the shape $3h\times 480 \times480$, where $h$ is the horizon. The tactile observation is processed by a tactile encoder consisting of a pre-trained ResNet-18 \cite{he2016deep} CNN (with the final classification layers removed), combined with a single convolutional layer (to map the $3h$ channels to 3 channels), resulting in a $\mathbb{R}^{3 \times h \times 480 \times 480} \to \mathbb{R}^{512}$ embedding.

To provide a horizon for the first few time steps, the BeadSight's tactile observation buffer is initialized with $h-1$ duplicates of the initial tactile frame. For these tasks, the sensor is not in contact with any object or surface at the start of each task run, so we would expect the initial frames to be nearly identical.

\subsubsection{Contrastive pre-training}
The vision and tactile encoders were pre-trained using the contrastive pre-training method described in \cite{george2024visuo}. This method leverages the relationships between the three different modalities of data our agent observes (visual, tactile, and positional) to train the encoders to extract task-relevant features. During pre-training, two projection networks are learned, one for visual observations and one for tactile and positional observations. These projection networks consist of an encoder (ResNet-18 for the image encoder and the BeadSight tactile encoder) along with a projection head that casts the encoded observations to a shared latent embedding space. The positional information is also passed to the tactile projection head. By adding position information to the tactile projector, we ensure the resulting latent space has both global information (position observations) and local information (tactile observations). 

To train the projection networks, a set of timesteps is sampled from a single trajectory, and the corresponding observations are projected to the shared latent space. Then, a CLIP loss is used to encourage the embeddings from the same timestep to be similar, while driving apart the projection of the observations from different timesteps. All of the camera views are encoded using a shared visual encoder. At each update, the contrastive loss is calculated separately for each camera and then summed, according to the equation: 

$$loss = -\sum_{c}\sum_{i \in t} \frac{1}{2n}(\log(\frac{\exp(sim_{i, i, c}/\tau)}{\sum_{j \in t} \exp(sim_{i, j, c}/\tau)})$$
$$ + \log(\frac{\exp(sim_{i, i, c}/\tau)}{\sum_{j \in t} \exp(sim_{j, i, c}/\tau)}))$$

Where $sim_{i, j, c}$ is the cosine similarity between the tactile embedding from timestep $i$ and the visual embedding of camera $c$'s observation from timestep $j$.

\subsubsection{Diffusion Policy}

Our Diffusion Policy approach leveraged \cite{chi2023diffusion}, and was modeled after \cite{george2024visuo}, with similar baseline hyper-parameters. We used the 1-D temporal CNN implementation, which was noted to be less sensitive to hyper-parameter tuning and to train stably for most tasks. Diffusion Policy is most effective when each camera view uses a separate trained encoder, so 6 ResNet-18 encoders were constructed and fine-tuned during training, one for each scene camera.  For the pretrained model, six copies of the pretrained ResNet-18 vision encoder were constructed and separately fine-tuned during training. 

After encoding, the vision and tactile observations were stacked to form a single observation vector, then passed to the network to condition its noise prediction. Our implementation uses an observation horizon of 1, and an action prediction horizon of 20. We also took advantage of noise scheduler decoupling, using 100 denoising steps during training for better resolution but using only 10 inference steps for faster execution. 

\subsubsection{Action Chunking Transformers}
We also used the same ACT architecture as \cite{george2024visuo}, with the only change being the replacement of the GelSight encoder with the BeadSight encoder. ACT calculates a set of goal actions based on the robot's current position, the latent encoding of the goal action sequence (set to zero during inference), and the encoded visual and tactile observations. Unlike Diffusion Policy, ACT does not apply an average adaptive pool to the output of the encoders, instead flattening the feature map to form an input sequence of shape (\textit{len}$\times 512$). As with \cite{george2024visuo}, we trained ACT on the relative position space, and at inference, implemented temporal ensembling in the global frame with temperature constant $k = 0.25$. 

\section{EXPERIMENTAL RESULTS}

\subsection{Cable Plugging Task}
\label{sec:cable results}

\begin{table}[b]
\caption{Success rates: USB cable plugging experiment}
\label{table:usb-results}
\begin{tabular}{|l|l|l|l|l|l|}
\hline
                                                                             & Diffusion & \begin{tabular}[c]{@{}l@{}}Diffusion \\ (Frozen)\end{tabular} & ACT  & \begin{tabular}[c]{@{}l@{}}Diffusion \\ (Gel)\end{tabular} & \begin{tabular}[c]{@{}l@{}}ACT\\ (Gel)\end{tabular} \\ \hline
\begin{tabular}[c]{@{}l@{}}Pre-trained, \\ Tactile + Vision\end{tabular}     & 0.05      & 0.8                                                           & 0.25 & 0.75                                                            & 0.95                                                     \\ \hline
\begin{tabular}[c]{@{}l@{}}Pre-trained, \\ Vision\end{tabular}               & 0.7       & -                                                             & 0.65 & 0.75                                                            & 0.85                                                     \\ \hline
\begin{tabular}[c]{@{}l@{}}No Pre-training, \\ Tactile + Vision\end{tabular} & 0.1       & 0.2                                                           & 0.5  & 0.7                                                             & 0.9                                                      \\ \hline
\begin{tabular}[c]{@{}l@{}}No Pre-training, \\ Vision\end{tabular}           & 0.05      & -                                                             & 0.15 & 0.45                                                            & 0.2                                                      \\ \hline
\end{tabular}
\end{table}

 To examine the effect of pre-training with a low-fidelity tactile sensor on imitation learning agents, we tested our trained ACT and Diffusion agents on the cable plugging task described in Section \ref{sec:data_collection}, following the same evaluation criteria used in \cite{george2024visuo} so that we can directly compare our results. For each evaluation, the agent attempted the task 20 times, with success being recorded if the agent plugged in and then released the USB. To increase the difficulty of the task, random noise was applied to the executed actions, sampled from a normal distribution with a standard deviation of 2.5 mm. 

For both ACT and Diffusion Policy, we evaluated the effect of low-fidelity visuo-tactile pre-training on both vision-only and visuo-tactile agents. Results are shown in Table \ref{table:usb-results} where ``Gel" denotes the GelSight sensor and all other results use the BeadSight sensor. ``Frozen" denotes freezing the tactile encoder (not updating the encoders weights) during downstream training.  

Prior work using GelSight demonstrated that visuo-tactile pre-training improves the performance of both vision-only and visuo-tactile agents, with the most notable gains observed in the vision-only setting. When replacing GelSight with BeadSight, an open-source, low-fidelity tactile sensor, we observed a shift in this trend: while pre-training with BeadSight negatively impacted \textbf{visuo-tactile} agents, reducing the success rate of both ACT and Diffusion Policy, its effect on \textbf{vision-only} agents was highly beneficial. Specifically, vision-only policies trained with BeadSight pre-training outperformed their visuo-tactile counterparts, demonstrating that even a low-cost sensor can still provide valuable training signals.

The gap between visuo-tactile and vision-only agents likely stems from over-fitting to BeadSight’s stochastic and variable observations. Unlike GelSight, BeadSight’s beads can drift significantly between experiments, leading to a train-test distribution mismatch. This explains why visuo-tactile agents struggled, particularly with Diffusion Policy, which lacks ACT’s temporal ensembling that mitigates out-of-distribution issues. This robustness is crucial in high-stakes tasks like cable plugging, where a single incorrect action can lead to irreversible failure.

To investigate BeadSight over-fitting, we compared its training and testing observations using a tSNE plot \cite{van2008visualizing}, generated from tactile embeddings extracted by the pre-trained encoder (Figure \ref{table:usb-results}). The results confirm a clear distribution shift between training and testing, highlighting the challenge of direct visuo-tactile policy transfer with BeadSight.

 \begin{figure}[h]
  \begin{center}
    \includegraphics[width=0.48\textwidth]{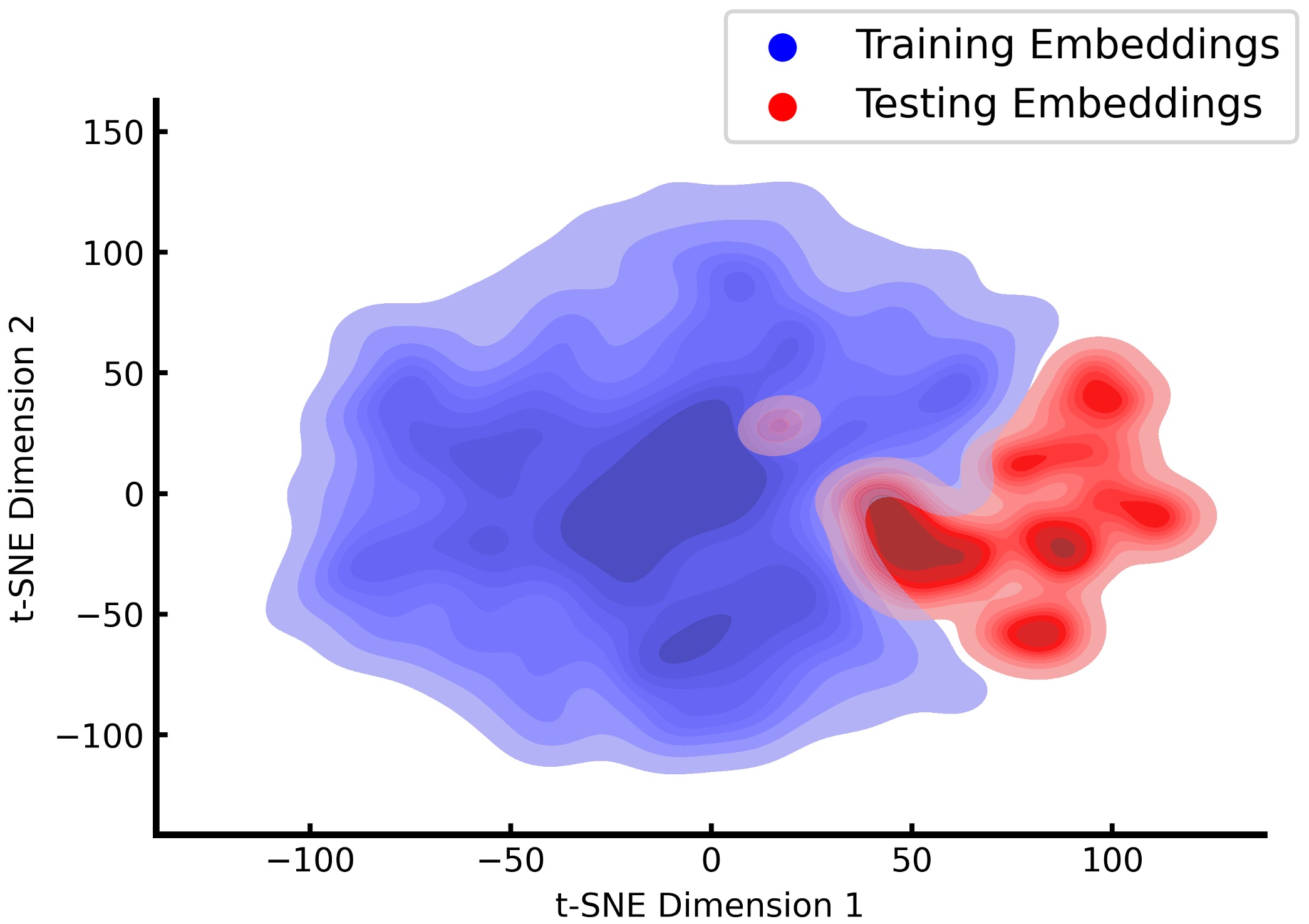}
  \end{center}
  \caption{tSNE plot showing the distribution shift between training and deployment of the tactile observation.}
  \label{fig:tsne}
\end{figure}

To mitigate this, we froze the BeadSight encoder weights before training the agents. For pre-trained models, we froze the weights obtained from contrastive pre-training; for non-pretrained models, we froze them at initialization. This approach effectively reduced over-fitting in Diffusion Policy, leading to substantial performance gains, surpassing both fine-tuned models and their respective vision-only and visuo-tactile baselines. However, due to ACT’s reliance on full feature maps rather than a compressed embedding, freezing the encoder was infeasible without major architectural modifications. Therefore, we only evaluated weight freezing on Diffusion Policy.

Comparing these results to those achieved in \cite{george2024visuo} using GelSight, we find that without pre-training, using a low-fidelity tactile sensor in imitation learning leads to a performance drop. However, when combined with visuo-tactile pre-training and weight freezing, the performance gap can be significantly reduced. Most importantly, when a vision-only policy is preferred, BeadSight offers a low-cost yet highly effective way to enhance learning, demonstrating that even open-source tactile sensors can provide meaningful benefits in policy training.

\subsection{Drawer Task}

\begin{table}[]
\caption{Success rates: drawer experiment, removed tactile encoder}
\label{table:drawer-results}
\begin{tabular}{|l|r|r|r|r|}
\hline
                          & \multicolumn{1}{l|}{Opened} & \multicolumn{1}{l|}{Picked} & \multicolumn{1}{l|}{Placed} & \multicolumn{1}{l|}{Closed} \\ \hline
No Pre-training           & 0.70                        & 0.35                        & 0.20                        & 0.20                        \\ \hline
Vision Pre-training  & 0.45                        & 0.25                        & 0.05                        & 0.05                        \\ \hline
Dissimilar Pre-training     & 0.90                        & 0.70                        & 0.50                        & .50               \\ \hline
Similar Pre-training & 1.00                        & 0.80                        & 0.50                        & .50               \\ \hline
Identical Pre-training    & 1.00                        & 0.90                        & 0.55                        & 0.55               \\ \hline
\end{tabular}
\end{table}

The drawer task evaluation consisted of the long-horizon drawer pick-and-place task. Each agent attempted the task 20 times. Success was assessed across the different sub-tasks (opening drawer, picking block, placing block in drawer, closing drawer). Each sub-task is only marked successful if the previous task was successful, meaning that the success rate for closing the drawer is also the total success rate for the overall task completion. Diffusion Policy was chosen for this task, and the effect of freezing vs. ablating the tactile encoder was compared. 

The supporting drawer task (similar) and original USB task (dissimilar) were used to investigate the effects of pre-training with different tasks. We also evaluated a vision pre-training ablation, which pre-trained without the tactile sensor to investigate whether the mere alignment of different camera views contributed to the performance improvement. For this ablation, all fixed camera images were aligned to the end-effector camera image and the end-effector position. This mimicked the visuo-tactile contrastive pre-training structure, where camera images were aligned to the tactile sensor images and end-effector position. 

The best-performing agent for this task removed, rather than froze, the tactile encoder before downstream training, as shown in Table \ref{table:drawer-results}. Both no pre-training and vision-only pre-training resulted in a 60–90\% performance decline compared to visuo-tactile pre-training across identical, similar, and dissimilar tasks. This underscores the essential role of tactile information in providing critical feedback for manipulation, and highlights that purely vision agents struggle without the rich contact information that tactile sensing offers. Furthermore, visuo-tactile pre-training on any of these tasks yielded comparable performance improvements, suggesting that a pre-trained visuo-tactile encoder can enhance any vision-only agent downstream, assuming the same robot system morphology.

The frozen tactile encoder agent (Table \ref{table:frozen}) produced mixed results, although the same technique led to an improvement for the previous cable plugging task. We speculate that the effect of freezing compared to ablating the tactile encoder depends on the drift of various physical parameters of the sensor, making this choice a hyper-parameter that can be tested to tune performance on different tasks. However, ablating the encoder is likely to be more stable, since as long as the sensor remains visually similar in the image, any drift on the tactile parameters will not affect the inference result. 

\begin{table}[]
\caption{Success rates: drawer experiment, frozen tactile encoder}
\label{table:frozen}
\begin{tabular}{|l|r|r|r|r|}
\hline
                          & \multicolumn{1}{l|}{Opened} & \multicolumn{1}{l|}{Picked} & \multicolumn{1}{l|}{Placed} & \multicolumn{1}{l|}{Closed} \\ \hline
Dissimilar Pre-training     & 0.45                        & 0.00                        & 0.00                        & 0.00              \\ \hline
Similar Pre-training & 0.95                        & 0.60                        & 0.50                        & 0.45             \\ \hline
Identical Pre-training    & 0.60                        & 0.35                        & 0.25                        & 0.20               \\ \hline
\end{tabular}
\end{table}

\section{CONCLUSION}
\label{sec:conclusion}

In this work, we explored combining a low-fidelity tactile sensor (BeadSight) with pre-training for imitation learning.  By using a low-fidelity, open-source tactile sensor and visuo-tactile pre-training, we were able to significantly improve the performance of a vision-only agent, replicating a prior finding that used a high-resolution GelSight tactile sensor. We show that freezing the tactile encoder after pre-training can be used to mitigate train-test distribution shifts when using a low-fidelity sensor, and the choice between freezing or ablating the sensor can act as a hyper-parameter (although ablating is likely more stable). 

Our follow-up investigation further supported this visuo-tactile pre-training technique. We showed that pre-training with vision alone could not lead to the same performance improvement and that visuo-tactile pre-training on a dissimilar task could benefit the downstream vision-only agent. 

These relative performance improvements used $\sim$ 100 expert demonstrations and low computational intensity imitation learning methods. This suggests tactile pre-training can be another tool for improving manipulation performance on difficult tactile-rich tasks with efficient computational methods. Other work in IL has attempted to dramatically reduce the number of demonstrations necessary for an agent to imitate an action \cite{parakh2023lifelong, george2023one, george2023minimizing} and one can imagine tactile pre-training as a way to further improve IL's efficiency on tactile-rich tasks. 

Future work could explore adding to the imitation learning pipeline proposed here. Currently, though the tactile sensor is not active during inference, it still needs to be physically attached to the end-effector to maintain consistent visual observations. Tracking and masking out the image of the tactile sensor could allow any gripper finger used in deployment, improving the utility of the vision-only agent. There is also the critical assumption of identical robot system morphology, which prevents learning across physically different robot platforms. This constraint could potentially be mitigated by exploring latent space pre-training approaches that leverage a larger visuo-tactile dataset, collected across a wide range of robot embodiments.

Finally, the benefit of freezing the encoder together with the pre-training approach implies a way to pre-train an encoder on a large, diverse dataset and use this to inform downstream tactile manipulation tasks, with only vision used at execution. This kind of pretrained model at an appropriate scale could broadly improve general robot manipulation.



\section*{APPENDIX}

We also include data from using a visuo-tactile agent with a fine-tuned tactile encoder for the drawer experiment in Table \ref{table:notfrozen}. 

In the original USB plugging experiment in Table \ref{table:usb-results}, this was denoted ``Tactile + Vision", where the agent used both vision and tactile observations, and the tactile encoder was fine-tuned when training the imitation learning agent (not frozen). Although this approach was already shown to perform worse than the vision-only agent when using BeadSight, we collected this data for the drawer experiment and included it for completion. Note that \cite{george2024visuo} demonstrated that with a more precise commercial sensor such as GelSight, the visuo-tactile agent out-performs the vision-only agent on the USB cable plugging task. 

\begin{table}[h]
\caption{Success rates: drawer experiment, finetuned tactile encoder}
\label{table:notfrozen}
\begin{tabular}{|l|r|r|r|r|}
\hline
                          & \multicolumn{1}{l|}{Opened} & \multicolumn{1}{l|}{Picked} & \multicolumn{1}{l|}{Placed} & \multicolumn{1}{l|}{Closed} \\ \hline
Dissimilar Pre-training Task     & 1.00                        & 0.55                        & 0.35                        & 0.30                        \\ \hline
Similar Pre-training Task & 0.85                        & 0.45                        & 0.10                        & 0.10                        \\ \hline
Identical Pre-training Task    & 0.75                        & 0.50                        & 0.25                        & 0.25                        \\ \hline
\end{tabular}
\end{table}



\bibliography{references}  




\end{document}